\documentclass[a4paper, 10 pt, conference]{ieeeconf}
\IEEEoverridecommandlockouts
\usepackage{epsfig}
\usepackage{amsmath}
\usepackage{array}
\usepackage{amssymb}
\usepackage{cite}
\usepackage{booktabs}
\usepackage{multirow}
\usepackage{subcaption}
\usepackage{booktabs}
\usepackage{mathtools}
\usepackage{siunitx}
\usepackage{comment}
\usepackage{threeparttable}
\usepackage{bm}
\usepackage{pifont}
\newcommand{\cmark}{\ding{51}}%
\makeatletter
\newcommand{\subsubsubsection}{\@startsection{paragraph}{4}{\z@}%
  {1.0\Cvs \@plus.5\Cdp \@minus.2\Cdp}%
  {.1\Cvs \@plus.3\Cdp}%
  {\reset@font\sffamily\normalsize}
}
\makeatother
\newcommand{\figlab}[1]{\label{fig:#1}}
\newcommand{\figref}[1]{Fig.~\ref{fig:#1}} 
\newcommand{\tablab}[1]{\label{tab:#1}}
\newcommand{\tabref}[1]{Table~\ref{tab:#1}} 

\newcommand{\algolab}[1]{\label{algorithm:#1}}
\newcommand{\algoref}[1]{Algorithm~\ref{algorithm:#1}} 

\newcommand{\etal}{\textit{et al.}}

\usepackage{color}
\usepackage{soul}
\newcommand{\kiyokawa}[1]{{{#1}}}

\usepackage{amsmath}
\usepackage{algorithmicx}
\usepackage[ruled]{algorithm}
\usepackage[noend]{algpseudocode}

\begin{document}
\title{Self-Supervised Learning of Grasping Arbitrary Objects On-the-Move}

\author{Takuya Kiyokawa$^{1,2}$, Eiki Nagata$^{1}$, Yoshihisa Tsurumine$^{1}$, Yuhwan Kwon$^{1}$, and Takamitsu Matsubara$^{1}$%
\thanks{$^{1}$Division of Information Science, Robot Learning Laboratory, Nara Institute of Science and Technology (NAIST), Ikoma, Nara, Japan.}%
\thanks{$^{2}$Department of Systems Innovation, Graduate School of Engineering Science, Osaka University, Toyonaka, Osaka, Japan.}
}

\maketitle

\begin{abstract}
\kiyokawa{\textit{Mobile grasping} enhances manipulation efficiency by utilizing robots' mobility. This study aims to enable a commercial off-the-shelf robot for mobile grasping, requiring precise timing and pose adjustments. Self-supervised learning can develop a generalizable policy to adjust the robot's velocity and determine grasp position and orientation based on the target object's shape and pose. Due to mobile grasping's complexity, action primitivization and step-by-step learning are crucial to avoid data sparsity in learning from trial and error. This study simplifies mobile grasping into two grasp action primitives and a moving action primitive, which can be operated with limited degrees of freedom for the manipulator. This study introduces three fully convolutional neural network (FCN) models to predict static grasp primitive, dynamic grasp primitive, and residual moving velocity error from visual inputs. A two-stage grasp learning approach facilitates seamless FCN model learning. The ablation study demonstrated that the proposed method achieved the highest grasping accuracy and pick-and-place efficiency. Furthermore, randomizing object shapes and environments in the simulation effectively achieved generalizable mobile grasping.}
\end{abstract}

\IEEEpeerreviewmaketitle

\section{Introduction}
To improve the efficiency of tidying the home environment using a mobile manipulator, grasping on-the-move (mobile grasping) is a promising approach. \figref{mobile_grasp} illustrates mobile grasping by a manipulator in both simulations (a) and the real world (b). Existing robotic mobile grasping methods for tidy-up tasks require stopping in front of each object to grasp them and are limited applicability to diverse objects~\cite{Taniguchi2021,Matsushima2022,Wu2023,Griffin2023}.

While many studies have successfully generated mobile grasping motions that account for uncertainties caused by the manipulator's movement~\cite{Colombo2019,Zimmermann2021,Limerick2023}, further efforts are needed to handle objects of various shapes.

Towards a practical application, this study aims to enable a commercial off-the-shelf robot to perform mobile grasping, which requires fine-tuning the timing and pose of the grasp.
A key feature of mobile grasping is that both the kinematics and dynamics of the moving hand affect the success rate. Thus, grasp plans for stationary objects cannot be directly applied. It is crucial to minimize movement errors that depend on the moving velocity and to execute grasping at the appropriate time. As it is challenging to manually teach these skills for grasping various shapes on the move, a framework enabling robots to self-acquire these abilities is necessary. Our approach utilizes the self-supervised learning method, which has previously shown high success rates in several manipulation tasks~\cite{Zeng2018,Zeng2020}.
\begin{figure}[tb]
    \centering
    \small
    \begin{minipage}[tb]{0.415\linewidth}
        \centering
        \includegraphics[keepaspectratio, width=\linewidth]{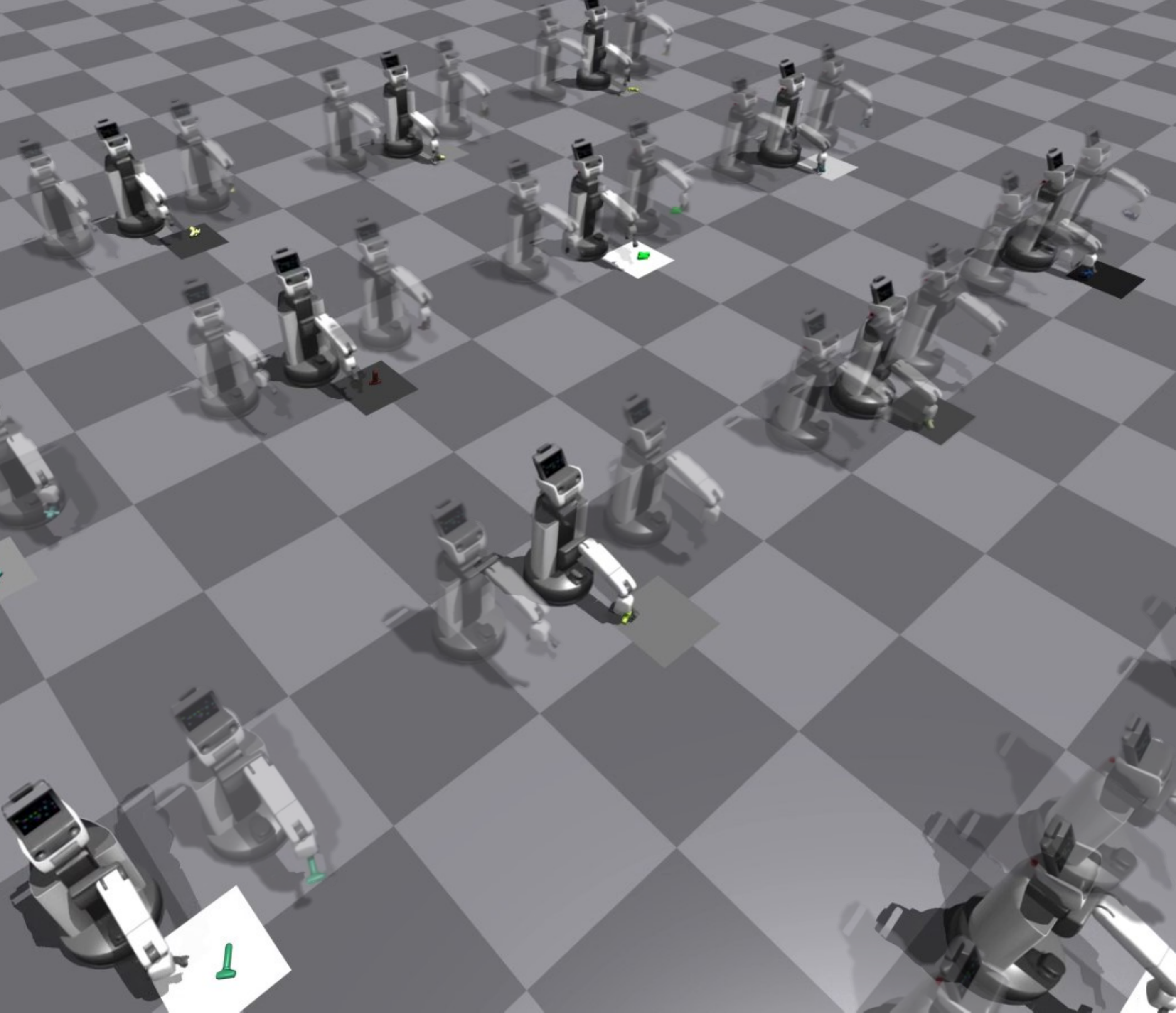}
        \subcaption{\small{Training in simulations}}
    \end{minipage}
    \begin{minipage}[tb]{0.57\linewidth}
        \centering
        \includegraphics[keepaspectratio, width=\linewidth]{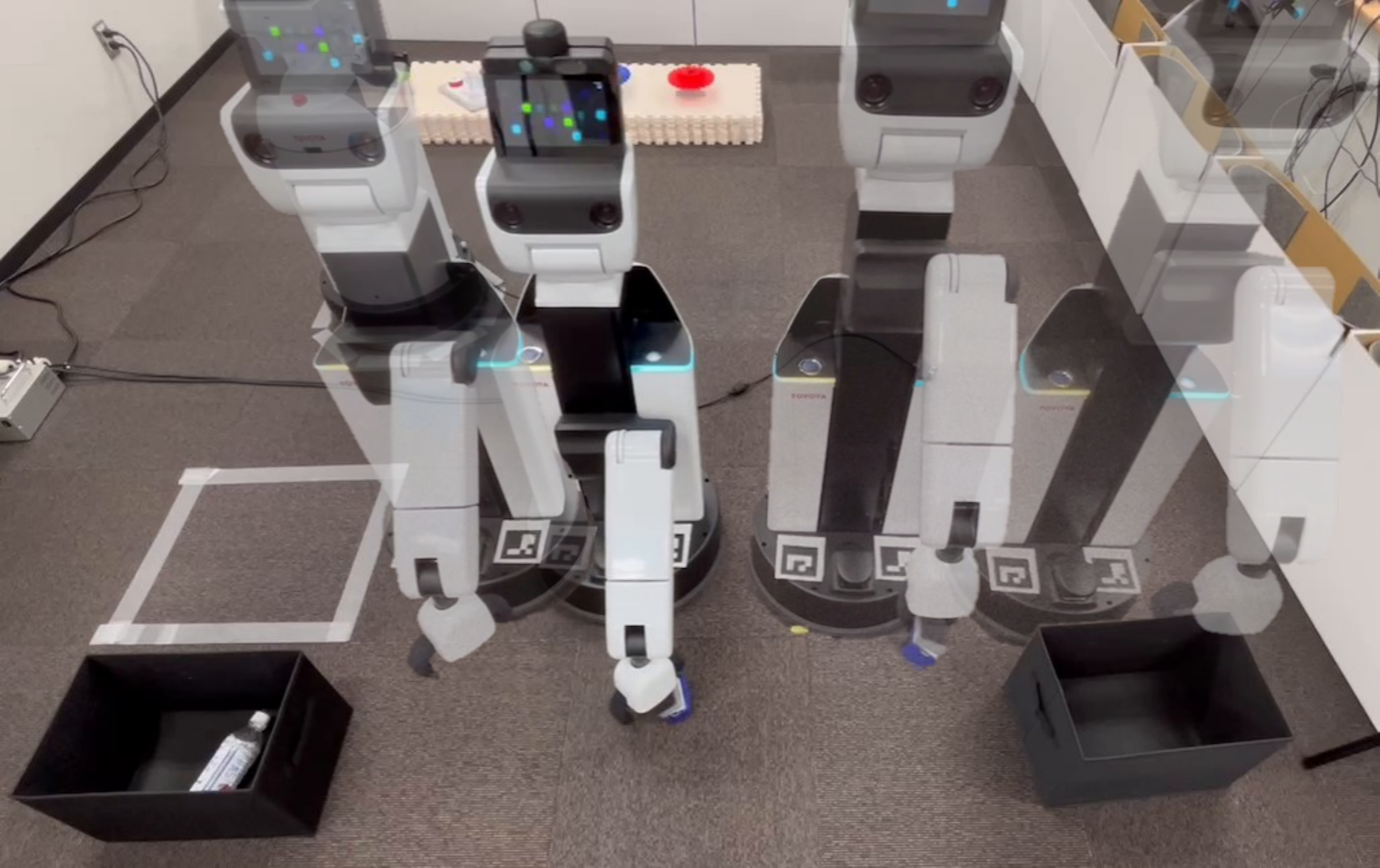}
        \subcaption{\small{Execution in the real world}}
    \end{minipage}
    \caption{\small{Scenes of mobile grasping for different objects in simulations and the real world.}}
    \figlab{mobile_grasp}
\end{figure}
\begin{figure}[tb]
    \centering
    \small
    \includegraphics[keepaspectratio, width=\linewidth]{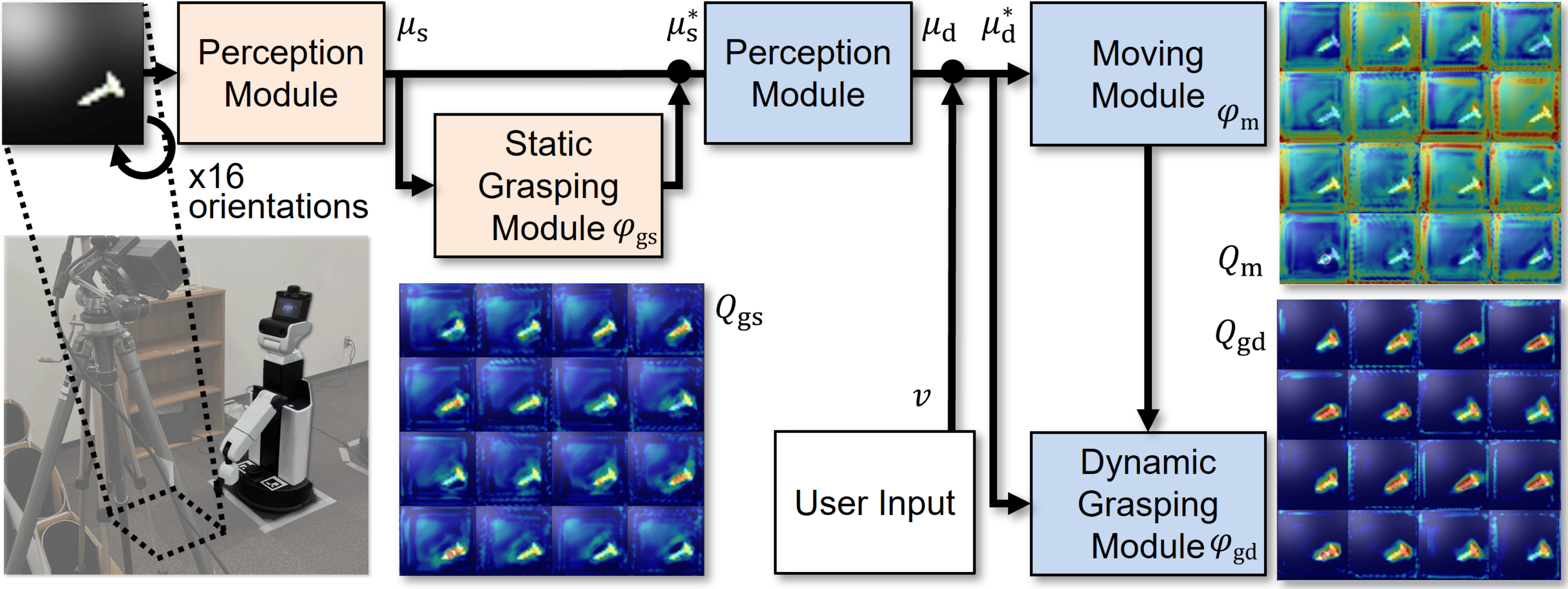}
    \caption{\small{Overview of the mobile grasping action generator. The models of the red boxes are first trained for the stationary object grasping action policy. These are then frozen, and the other models of the blue boxes are trained to learn the moving and dynamic grasping action policies.}}
    \figlab{overview}
\end{figure}

\figref{overview} illustrates the proposed mobile grasping action generator, comprising three modules utilizing fully convolutional networks (FCNs) tailored to specific abilities. The self-supervised learning framework allows models to train through trial and error using robot-generated correct labels.

To address the inefficiency of mobile manipulators learning grasping from scratch, we propose a dynamic grasping module leveraging a static grasping module pretrained for stationary objects. This yields a two-stage learning process where static grasping serves as the foundation for acquiring mobile grasping skills. Both grasping modules assume a parallel two-finger gripper approaches the target vertically to pick it up. These modules estimate pixel-wise grasping success probabilities, generalizing to mobile grasping of various shapes.

To mitigate grasp errors caused by discrepancies between actual and desired moving velocities, we introduce an FCN model that predicts residual velocity, adjusting commanded velocity to enhance grasping success. Essentially, this study demonstrates the efficacy of the proposed models in predicting dynamic grasp success probability based on static grasp predictions and residual velocity. Each model specializes in its role, facilitating efficient learning. During training, the moving velocity and target object shape are varied randomly to develop a model adaptable to diverse velocities and shapes.

The major contributions of this work are twofold:
\begin{enumerate}
    \item \kiyokawa{We propose a self-supervised learning framework employing a two-stage grasp learning method to integrate three action primitive-specific FCN models generalized by randomization techniques, achieving object-agnostic mobile grasping ability.}
    \item \kiyokawa{The method is widely applicable to practical applications, using only a commercial off-the-shelf manipulator with limited degrees of freedom (DoFs) and RGB-D sensor, which facilitates setup and maintenance. Our mobile grasping system empirically show higher success rate and efficiency through thorough simulations and real-world tests.}
\end{enumerate}

Our experiments involved mobile grasping at different velocities for target objects of various shapes, demonstrating the effectiveness of the proposed method by comparing grasping success rates, learning performance, and tidy-up task completion times against methods excluding each module, in both simulations and real-world scenarios.

\section{Related Work}
\subsection{Robotic Tidy-Up Tasks}
\kiyokawa{Previous robotic tidy-up methods for home environments primarily focused on enhancing picking accuracy, involving the sequential stopping and grasping of objects in static environments}~\cite{Abdo2015,Griffin2023,Wu2023}. 
Abdo~\etal~\cite{Abdo2015}\kiyokawa{ enabled robots to predict user preferences for organizing objects in containers like shelves or boxes.} Griffin~\etal~\cite{Griffin2023}\kiyokawa{ introduced a detection-based manipulation method and an efficient learning approach for detectors.} Wu~\etal~\cite{Wu2023}\kiyokawa{ employed large language models to achieve personalized tidy-up tasks based on user preferences. However, existing tidy-up robots stop in front of objects before grasping them.}

\kiyokawa{This study seeks to improve tidy-up task efficiency by enabling robots to grasp objects while moving, potentially reducing task duration significantly. It does not address high-level planning for scheduling tidy-up tasks, as studied in}~\cite{Yan2021,Su2023}\kiyokawa{, which is crucial for minimizing overall operation time.}

\subsection{Self-Supervised Grasp Learning}
\kiyokawa{Untrained objects generally require recognition models to be retrained with new datasets. Data-driven methods}~\cite{Redmon2015,Pinto2015,Lenz2015} \kiyokawa{have shown that deep learning (DL) models trained on grasp detection datasets with diverse objects and multiple labeled grasps can eliminate the need for 3D models. This approach is scalable to tasks like grasping, pushing, and picking}~\cite{Pinto2016}. Zeng~\etal~\cite{Zeng2018icra} \kiyokawa{improved bounding box-based grasp detection by creating an object-agnostic framework that maps visual observations to actions and infers dense pixel-wise affordance maps for various grasping actions.}

\kiyokawa{A leading solution involves a model that outputs pixel-wise grasp probabilities from visual data}~\cite{Zeng2018,Zeng2020}. \kiyokawa{Similar to}~\cite{Zeng2018icra}, \kiyokawa{these methods use FCNs to deduce pixel-wise affordances on RGB-D images and apply Q-learning for self-supervised learning through trial and error. This method achieves high success rates and efficiency in grasping after several hours of training and adapts to new objects.
This study is a first attempt to challenge the pixel-wise self-supervised learning for mobile grasping of diverse objects.}

\subsection{\kiyokawa{Dynamic Grasping}}
\kiyokawa{Mobile grasping can be regarded as a dynamic grasping method.
Studies have explored dynamic grasping for various applications, such as grasping moving objects on conveyor belts}~\cite{Akinola2021} \kiyokawa{and tracking and grasping stationary objects using a moving camera}~\cite{Huang2023}. Akinola~\etal~\cite{Akinola2021} \kiyokawa{and} Huang~\etal~\cite{Huang2023} \kiyokawa{did not employ mobile robots, focusing instead on other dynamic tasks.}

\kiyokawa{Mobile grasping with non-mobile manipulators, like drones}~\cite{Fishman2021} \kiyokawa{and quadruped robots}~\cite{Zimmermann2021}\kiyokawa{, has been achieved for simple object shapes.
These scenarios present distinct challenges due to the varied uncertainties introduced by different robots' DoFs.}

\kiyokawa{Mobile manipulators can execute interactive tasks in diverse environments, including industrial, domestic, and natural settings. Previous research has explored mobile grasping with mobile manipulators}~\cite{Colombo2019,Thakar2022tase,Limerick2023}.
Colombo~\etal~\cite{Colombo2019}\kiyokawa{ developed a nonlinear model predictive controller for precise trajectory tracking in mobile grasping, requiring fewer online computations.}
Thakar~\etal~\cite{Thakar2022tase}\kiyokawa{ introduced a bidirectional sampling-based scheme for generating trajectories for mobile base and grasping poses to reduce operation time in large spaces.}
Limerick~\etal~\cite{Limerick2023}\kiyokawa{ designed a mobile manipulation architecture with continuous base motion, capable of adapting to environmental changes and grasping unpredictably moving objects while in motion.}
\kiyokawa{However, these approaches have not been validated for varying object shapes and are limited to known shapes}~\cite{Colombo2019,Thakar2022tase,Limerick2023}.

\kiyokawa{This study tackles the problem of grasping arbitrary objects in motion within home environments. Self-supervised learning is employed to enable mobile manipulators to grasp unknown objects.
Compared to previous studies on mobile grasping with mobile manipulators}~\cite{Colombo2019,Thakar2022tase,Limerick2023}\kiyokawa{, this study assesses the proposed method through simulations and real-world experiments involving actual household objects of various shapes.}

\section{Proposed Method}
\begin{figure}[tb]
    \centering
    \small
    \includegraphics[keepaspectratio, width=0.82\linewidth]{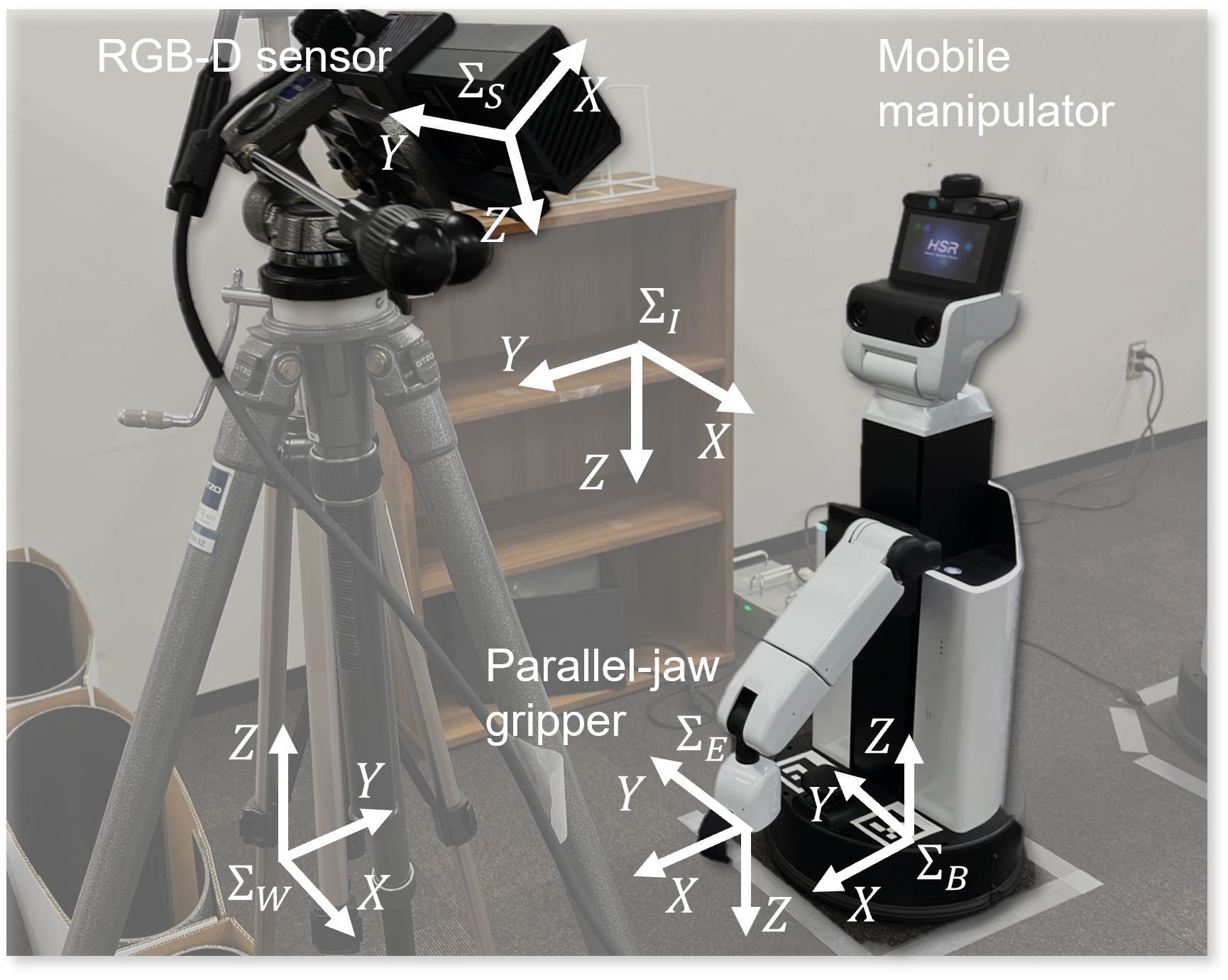}
    \caption{\small{Coordinate systems defined in the experimental environment}}
    \figlab{cs}
\end{figure}
\subsection{Overview}
\alglanguage{pseudocode}
\begin{algorithm}[t]
\caption{Mobile Grasping Action Generation} \algolab{pipeline}
\begin{algorithmic}[1]
\renewcommand{\algorithmicrequire}{\textbf{Input:}}
\renewcommand{\algorithmicensure}{\textbf{Output:}}
\Require RGB-D heightmap image $I$, User input velocity $v$
\Ensure Action primitives $\varphi_{\rm{m}}$ and $\varphi_{\rm{gd}}$
\State $\mu_{\rm{s}}$ $\leftarrow$ PerceptionModule1($I$)
\State $Q_{\rm{gs}}$ $\leftarrow$ StaticGraspingModule($\mu_{\rm{s}}$)
\State $\mu_{\rm{s}}^*$ $\leftarrow$ Concatenation1($\mu_{\rm{s}}$, $Q_{\rm{gs}}$)
\State $\mu_{\rm{d}}$ $\leftarrow$ PerceptionModule2($\mu_{\rm{s}}$, $Q_{\rm{gs}}$)
\State $\mu_{\rm{d}}^*$ $\leftarrow$ Concatenation2($v$, $\mu_{\rm{d}}$)
\State $Q_{\rm{m}}$ $\leftarrow$ MovingModule($\mu_{\rm{d}}^*$) 
\State $Q_{\rm{gd}}$  $\leftarrow$ DynamicGraspingModule($\mu_{\rm{d}}^*$, $Q_{\rm{m}}$)
\State $\varphi_{\rm{m}}$, $\varphi_{\rm{gd}}$ $\leftarrow$ ActionDetermination($Q_{\rm{m}}$, $Q_{\rm{gd}}$)
\State \textbf{return} $\varphi_{\rm{m}}$, $\varphi_{\rm{gd}}$
\end{algorithmic}
\end{algorithm}
\figref{overview} shows the outline of the proposed generator and \algoref{pipeline} outlines the algorithm to show the flow of processes.
An RGB-D sensor fixed at a height higher than that of the robot, and looking down on the floor, provides the input of the generator.
Convert the obtained RGB-D image to a point cloud and generate a heightmap $I$ the pixel values of which are the heights from the floor, as follows:
Based on the generator’s predictions for the input image, which includes the entire outline of the target object, the robot moves in a straight line toward the target object and grasps it.

The generator has models for perceptual processing, static grasping, moving, and dynamic grasping modules composed of 7-layer fully convolutional residual networks~\cite{He2016}.
The input is a heightmap $I$ (112$\times$112$\times$4$\times$16) rotated by multiples of 22.5 degrees (16 rotations) to account for different gripper orientations.
The input image is converted into a visual feature $\mu_{\rm{s}}$ (40$\times$40$\times$512$\times$16) using the preceding perception module, PerceptionModule1().

We introduce the perception module with reference to the method of a previous study~\cite{Zeng2020}, which has shown high performance in self-supervised learning of multiple types of action policies based on visual information.
The static grasping module, StaticGraspingModule(), generates a static grasp primitive $\varphi_{\rm{gs}}$ for stationary objects based on the visual information.
$\mu_{\rm{s}}$ is fed into the static grasping module and concatenated with the static grasp success probability map, $Q_{\rm{gs}}$ which is the output of the static grasping module (Concatenation1()).
The concatenated visual feature $\mu_{\rm{s}}^*$ is fed into the succeeding perception module, PerceptionModule2().

Learning of both the perceptual processing module that outputs $\mu_{\rm{s}}$ and the static grasping module is performed before starting to train the moving and dynamic grasping modules.
In other words, in the phase to start training the models of the moving and dynamic grasping modules, the models of the static grasping and preceding perception modules are assumed to be already trained and frozen.
\figref{overview} shows the frozen models and the other models to be trained as red and blue boxes, respectively.

The moving velocity vector (40$\times$40$ \times$128$\times$16) generated based on the user input is concatenated with the visual feature $\mu_{\rm{d}}$ (40$\times$40$\times$512$\times$16) from the succeeding perception module as the final channel of $\mu_{\rm{d}}$ (Concatenation2()).
The concatenated visual feature $\mu_{\rm{d}}^*$ (40$\times$40$\times$640$\times$16) is fed into the moving and dynamic grasping modules, DynamicGraspingModule().

Subsequently, the generator computes the final moving velocity by predicting the residual velocity.
The dynamic grasping module needs to make inferences by considering the final moving velocity, which takes into account the residual velocity; therefore, the residual velocity map $Q_{\rm{m}}$ (112$\times$112$\times$1$\times$16) which is obtained by MovingModule() with $\mu_{\rm{d}}^*$ is also input for the dynamic grasping module.

Finally, $Q_{\rm{m}}$ and the dynamic grasp success probability map $Q_{\rm{gd}}$ (112$\times$112$\times$1$\times$16) are outputted, and the mobile manipulator performs the generated mobile grasping action based on the action primitives $\varphi_{\rm{m}}$ and $\varphi_{\rm{gd}}$ determined by the method described in the following section (ActionDetermination()).
This study uses a two-stage grasp learning method. The static grasping module facilitates the learning of the mobile grasping task execution policy by starting from an easier task setting, and then the dynamic grasping module seamlessly proceeds with learning of the final grasp action while benefiting from the static grasp inference, taking the final moving velocity as an input.

\subsection{Motion Primitives Defined for a Mobile Grasping Action}
Variables with several parameters, $\varphi_{\rm{gd}}=(x,y,\theta)$ and $\varphi_{\rm{m}}=\hat{v}$ define the mobile grasping action.
\figref{cs} draws the world coordinate system $\Sigma_W$, robot base coordinate system $\Sigma_B$, end-effector coordinate system $\Sigma_E$, sensor coordinate system $\Sigma_S$, and the image coordinate system $\Sigma_I$.
$\Sigma_S$ and $\Sigma_I$ are fixed coordinate systems, $\Sigma_I$ is a coordinate system for transforming the captured image to the image from directly above, and the coordinate values can be transformed between them using a predetermined homographic transformation matrix.
$\Sigma_W$ and $\Sigma_B$ are also fixed and the four $\Sigma_W$, $\Sigma_B$, $\Sigma_S$, and $\Sigma_I$ are calibrated in advance.
The coordinate values can be converted into each other.

Based on the parameters of the dynamic grasp primitive, $\varphi_{\rm{gd}}$, the mobile grasping action is generated as follows:
$y$ is the value in the $Y$-axis direction (direction perpendicular to the moving direction) of $\Sigma_W$ calculated based on the position of the highest pixel value of $Q_{\rm{gd}}$.
The initial position of the robot is adjusted by $y$.
$x$ corresponds to the coordinate position on the $X$-axis of $\Sigma_W$, which is the direction of movement.
$\theta$ represents the rotation angle around the $Z$-axis of $\Sigma_B$, which is the orientation of the end-effector approaching from directly above the target object.

The image containing the pixel with the highest score in the 16-dimensional images of $Q_{\rm{gs}}$ determines the rotation angle indicated by that dimension as $\theta$ as the orientation angle.
Parameter $\hat{v}$ of the moving primitive $\varphi_{\rm{m}}$ is the final moving velocity adjusted by adding the predicted residual velocity to the user input $v$.
Here, the pixel value of $Q_{\rm{m}}$ at the position where the score of the pixels in $Q_{\rm{gd}}$ is the highest is regarded as the predicted value of the residual velocity.
This is the actual movement speed corrected by adding the pixel value of the residual velocity to the user input $v$.
The velocity is controlled to be $\hat{v}$ after initializing the mobile base start position, and according to $x$ and $\theta$, the grasp position and orientation are adjusted to perform the generated mobile grasping action.

\subsection{Self-Supervised Mobile Grasping Learning}
At each time step in the self-supervised learning of the mobile grasping task execution policy, an RGB-D image is fed into the pipeline, and a heightmap $I$ is generated.
To generalize the models to a variety of moving velocities, $v$ is randomly assigned within the range of 0.10–0.20~[m/s].
To build the static grasping module, we use a model that learns only the grasping module proposed in~\cite{Zeng2018}.
The weights of the FCNs are frozen and the moving and dynamic grasping modules are trained after learning.

Based on the motion primitives $\varphi_{\rm{m}}$ and $\varphi_{\rm{gd}}$ determined from the output of each module, a mobile grasping action is generated and executed.
After execution, the success of the grasp is determined by checking whether the end-effector is fully closed after the grasping execution.
If the end-effector is not closed, the grasp is determined to be successful.
This enables self-supervised learning because the robot itself can determine the grasping successes and failures.
During learning, the model weights are updated based on the loss function by obtaining the binary correct label of the grasp $\mathcal{G}_i\in\{0,1\}$ at pixel $i$ of input $I$ and the correct residual velocity value $\bar{\delta}_i$, which is calculated from the ground truth of the moving velocity at the successful grasp and the target moving velocity given at that time.

The loss function used for learning is $\mathcal{L}=\mathcal{L}_g+\mathcal{G}_i\mathcal{L}_m$, and $\mathcal{L}_g$ is the Binary Cross Entropy Loss represented as the following equation:
\begin{equation}
    \begin{split}
        \mathcal{L}_g = - (\mathcal{G}_i \log q_i + (1-\mathcal{G}_i) \log (1-q_i)),
    \end{split}
\end{equation}
where $q_i$ represents the predicted value of grasping success rate corresponding to the image pixel of the output of dynamic grasping module $Q_{\rm{gd}}$.
$\mathcal{L}_m$ denotes Huber Loss value regarding the residual velocity as follows:
\begin{equation}
    \begin{aligned}
        \mathcal{L}_m = 
            \begin{cases}
            \frac{1}{2}(\delta_i - \bar{\delta}_i)^2, & \text{for } \mathcal{G}_i = 1 \text{ and } |\delta_i - \bar{\delta}_i| < 1, \\
            0, & \text{for } \mathcal{G}_i = 0, \\
            |\delta_i - \bar{\delta}_i| - \frac{1}{2}, & \text{otherwise,}
            \end{cases}
    \end{aligned}
\end{equation}
where $\delta_i$ represents the predicted value of the residual velocity corresponding to the pixel value of $Q_{\rm{m}}$ in the model output of the moving module.
Using this equation, $\mathcal{L}$ is calculated, and the weights of the models of the moving and dynamic grasping modules are updated by the back propagation of the FCN models.

\kiyokawa{To generalize the ability to manipulate and grasp objects of various appearances and shapes, we utilized target objects randomly colored and shaped by combining multiple rectangular forms, thus broadening the grasping capability}~\cite{Tobin2018}.
\kiyokawa{Similarly, to generalize the models to different backgrounds, we employed randomly colored workspaces. The moving and dynamic grasping modules were trained through 500 steps of trial and error in each of the 16 workspaces and tested in varied environments with different velocity settings, further tested in 1000 trials per velocity setting. Besides specific velocity settings of $0.10$, $0.15$, and $0.20$, another case used a randomly selected velocity setting $v^{rv}$ from the set $\{0.10, 0.11, \ldots, 0.22\}$.}

\section{Training and Evaluating Models}
\begin{figure}[tb]
    \centering
    \small
    \begin{minipage}[tb]{0.49\linewidth}
        \centering
        \includegraphics[keepaspectratio, width=\linewidth]{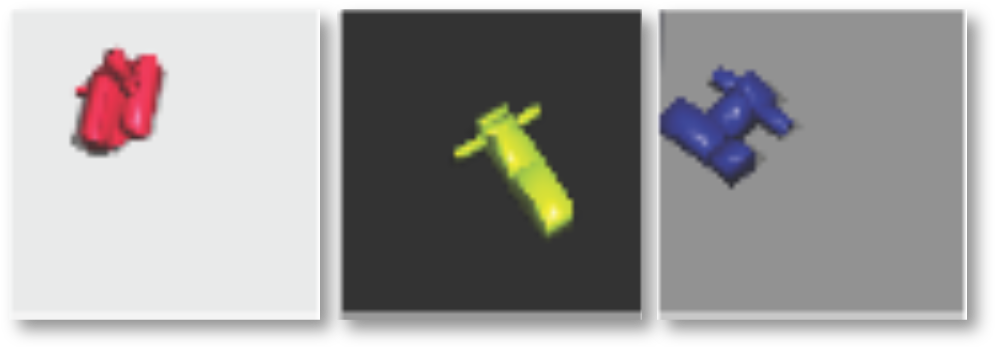}
        \subcaption{\small{Simulation}}
    \end{minipage}
    \begin{minipage}[tb]{0.49\linewidth}
        \centering
        \includegraphics[keepaspectratio, width=\linewidth]{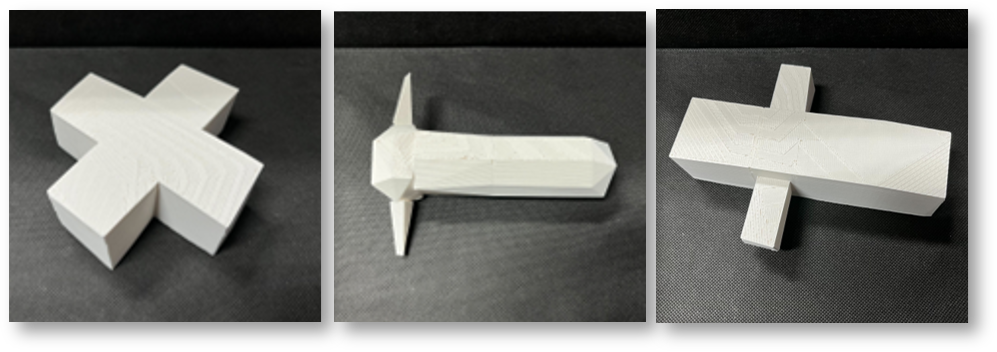}
        \subcaption{\small{Real-world}}
    \end{minipage}
    \caption{\small{Target objects used for (a) training and testing models in simulations and (b) finetuning the models in the real world.}}
    \figlab{obj}
\end{figure}
To evaluate the effectiveness of the proposed framework, we investigated the trained models and their performance in terms of mobile grasping success rate.
\figref{cs} shows the hardware devices. \figref{obj}~(a) shows target objects used in our simulation experiments.
We used a simulator equipped with a physics engine that performs mobile grasping using a mobile manipulator, the Toyota Human Support Robot (HSR), and evaluated the effectiveness of the proposed method by comparing it with other methods that remove certain components.
To quickly finish the training, we constructed a training environment in which multiple HSRs using NVIDIA Isaac Gym perform mobile grasping actions in parallel.
Multiple agents perform mobile grasping actions on a target object in the workspace in front of the mobile manipulator.

\subsection{Comparative Methods}
\begin{table}[tb]
    \centering
    \small
    \begin{threeparttable}
        \caption{\small{Differences among mobile grasping methods used in our experiments}}
        \tablab{comp_methods}
        \begin{tabular}{cccccc} \toprule 
            & BL$^{\rm *a}$ & w/o SG$^{\rm *b}$ & w/o M$^{\rm *c}$ & w/o DG$^{\rm *d}$ & PP$^{\rm *e}$  \\ \midrule
            SG & \cmark &  & \cmark & \cmark & \cmark \\
            M &  & \cmark &  & \cmark & \cmark \\
            DG &  & \cmark & \cmark &  & \cmark \\ \bottomrule
        \end{tabular}
        \begin{tablenotes}
            \item[*a]\footnotesize{Baseline method}
            \item[*b]\footnotesize{Proposed method without Static Grasping Module}
            \item[*c]\footnotesize{Proposed method without Moving Module}
            \item[*d]\footnotesize{Proposed method without Dynamic Grasping Module}
            \item[*e]\footnotesize{Proposed method}
        \end{tablenotes}
    \end{threeparttable}
\end{table}
To clarify the performance of each module in the proposed method, we prepared four comparative methods that removed one module.
\tabref{comp_methods} shows the difference between the four comparative methods and the proposed method (PP).
The comparative methods included the reference method (BL), w/o static grasping module (w/o SG), w/o moving module (w/o M), and w/o dynamic grasping module (w/o DG).

\textit{\textbf{BL.}} This approach removes the moving and dynamic grasping modules from the PP and executes the grasp predicted by the static grasping module.
This method performs the mobile grasp without the ability to predict errors in the moving velocity, and without the grasping ability specific to the dynamic grasp.

\textit{\textbf{w/o SG.}} This approach is regarded as the PP without the static grasping module.
This method learns the mobile grasping ability from a scratch.
Unlike the PP, this method cannot predict dynamic grasp with reference to the output of the static grasping module.

\textit{\textbf{w/o M.}} This approach does not explicitly learn the residual velocity to adjust the actual moving velocity, which implies that the moving module is removed from the PP.
This approach learns only static and dynamic grasping abilities, and maintains the moving velocity at the speed given by the user.
In other words, the PP requires the dynamic grasping module to acquire the ability to grasp with the error of the moving velocity by itself based on the moving velocity and image features, which places a heavy burden on a single module.
PP eliminates this burden by dividing the workload between multiple modules, allowing each module to focus on its own role.

\textit{\textbf{w/o DG.}} This approach removed the dynamic grasping module from the PP.
The grasping position generated by the static grasping module was used, and only the moving module was responsible for resolving complex uncertainties in the mobile grasp.
This method relyies on that the moving module is forced to absorb all the differences between the static and dynamic grasps.

\subsection{Grasping Success Rate Comparisons}
\begin{table}[tb]
    \centering
    \small
    \begin{threeparttable}
        \caption{\small{Mobile grasping success rate in simulations [\%]}}
        \tablab{res_v}
        \begin{tabular}{cp{8mm}p{8mm}p{8mm}p{8mm}p{8mm}} \toprule
            \multicolumn{1}{c}{$v$[m/s]} & \multicolumn{1}{c}{~~~BL~~~} & \multicolumn{1}{c}{~w/o SG~} & \multicolumn{1}{c}{~w/o M~} & \multicolumn{1}{c}{~w/o DG~} & \multicolumn{1}{c}{~~~PP~~~} \\ \midrule
            $v^{rv}$ & \multicolumn{1}{r}{73} & \multicolumn{1}{r}{79} & \multicolumn{1}{r}{80} & \multicolumn{1}{r}{78} & \multicolumn{1}{r}{\textbf{83}} \\
            0.10 & \multicolumn{1}{r}{74} & \multicolumn{1}{r}{82} & \multicolumn{1}{r}{83} & \multicolumn{1}{r}{79} & \multicolumn{1}{r}{\textbf{84}} \\
            0.15 & \multicolumn{1}{r}{71} & \multicolumn{1}{r}{82} & \multicolumn{1}{r}{82} & \multicolumn{1}{r}{78} & \multicolumn{1}{r}{\textbf{83}} \\
            0.20 & \multicolumn{1}{r}{69} & \multicolumn{1}{r}{79} & \multicolumn{1}{r}{77} & \multicolumn{1}{r}{76} & \multicolumn{1}{r}{\textbf{83}} \\ \midrule
            Mean & \multicolumn{1}{r}{72}	& \multicolumn{1}{r}{81} & \multicolumn{1}{r}{81} & \multicolumn{1}{r}{78} & \multicolumn{1}{r}{\textbf{83}} \\ \bottomrule
        \end{tabular}
    \end{threeparttable}
\end{table}
\begin{figure}[tb]
    \centering
    \small
    \includegraphics[keepaspectratio, width=0.8\linewidth]{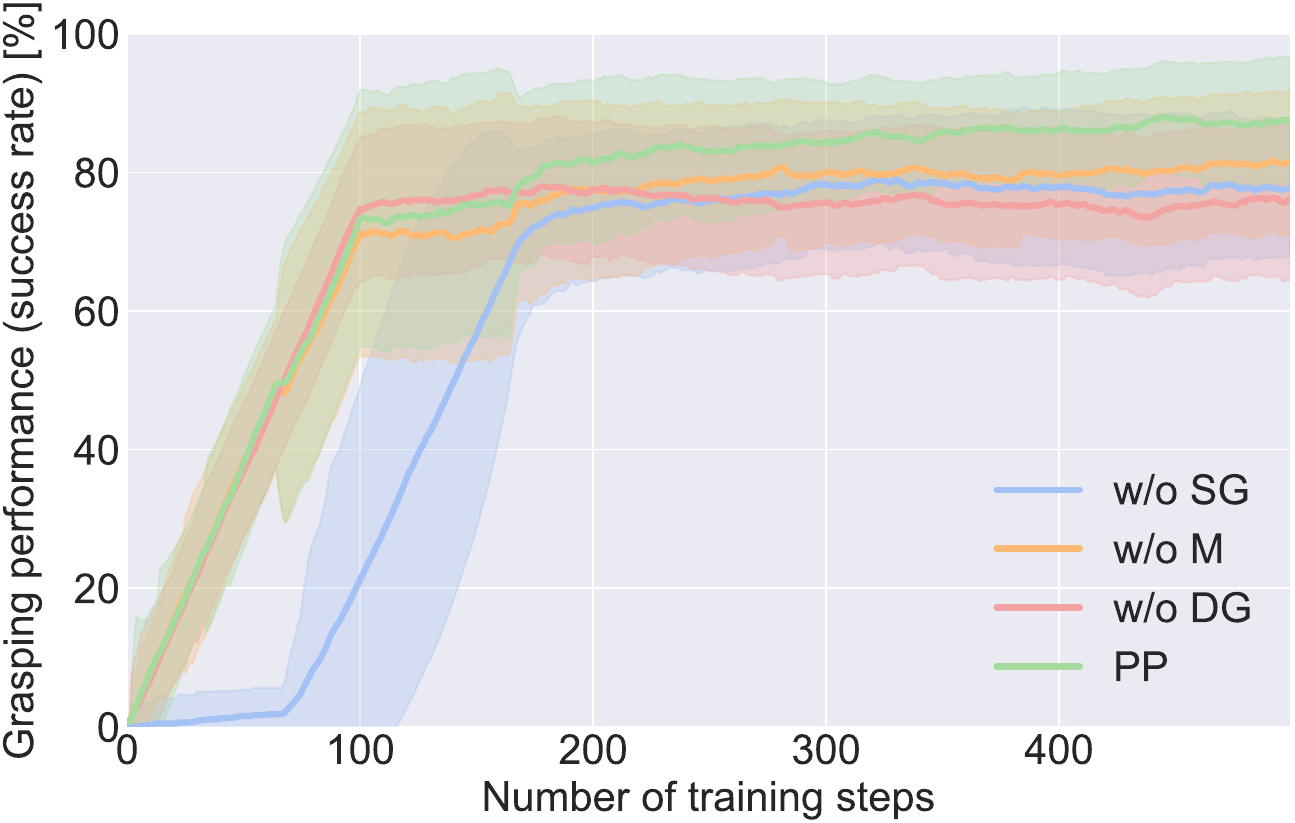}
    \caption{\small{Learning curve representing the relationship between the grasping success rate and increasing training steps}}
    \figlab{training_curve}
\end{figure}
\tabref{res_v} shows the grasping success rate. For $v^{rv}$, the PP was the highest.
Looking at other comparative methods, the lower the velocity, the higher the mobile grasping performance.
This result suggests that the higher the moving velocity, the more difficult the mobile grasping and the lower the success rate.
Furthermore, when comparing the results of the PP with those of the comparative methods, the PP was the highest at all velocities compared to the other methods.
Compared with the lowest success rates of all comparative methods, that is, $v^{rv}, 0.10, 0.15$, and $0.20$, the PP improved by 9.3\%, 9.9\%, 11.9\%, and 14.2\%, respectively.

\kiyokawa{While the PP enhances mobile grasping success rates, several failures still occur when attempting to grasp thin objects with limited protrusions suitable for two-finger grasping.} Morrison~\etal~\cite{Morrison2020} \kiyokawa{demonstrated that antipodal grasps for these shapes are challenging, indicating the potential need for alternative grippers in mobile grasping methods.}

\figref{training_curve} shows the mean values of grasping success rates (grasp performance) [\%] of all comparative methods except the BL.
We calculated grasp performance as the success rate of the last 1000 grasps generated with the model during the training step.
For w/o M, w/o DG, and PP, which are methods that utilize the output of the static grasping module, the success rates reached 70\% by the 100th step, indicating that the proposed learning method proceeded efficiently owing to the effect of the static grasping module.

\subsection{Learning Stable Mobile Grasping}
\begin{figure}[tb]
    \centering
    \small
    \begin{minipage}[tb]{0.49\linewidth}
        \centering
        \includegraphics[keepaspectratio, width=\linewidth]{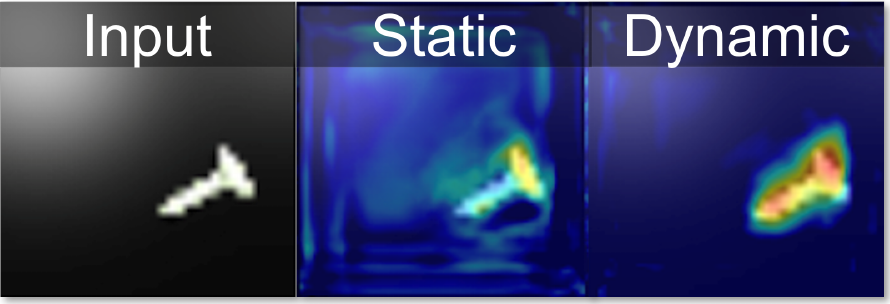}
        \subcaption{\small{T-shaped object 1}}
        \vspace{1mm}
    \end{minipage}
    \begin{minipage}[tb]{0.49\linewidth}
        \centering
        \includegraphics[keepaspectratio, width=\linewidth]{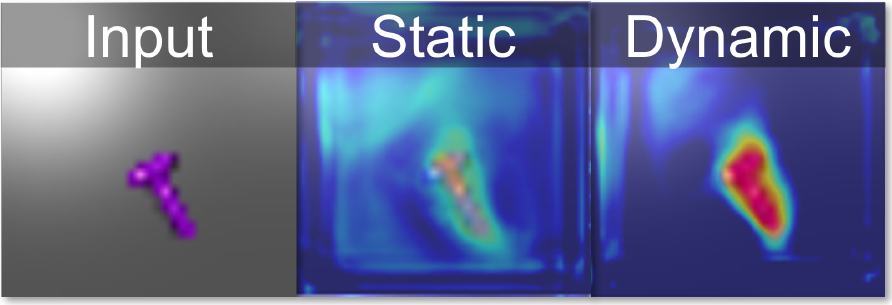}
        \subcaption{\small{T-shaped object 2}}
        \vspace{1mm}
    \end{minipage}
    \begin{minipage}[tb]{0.49\linewidth}
        \centering
        \includegraphics[keepaspectratio, width=\linewidth]{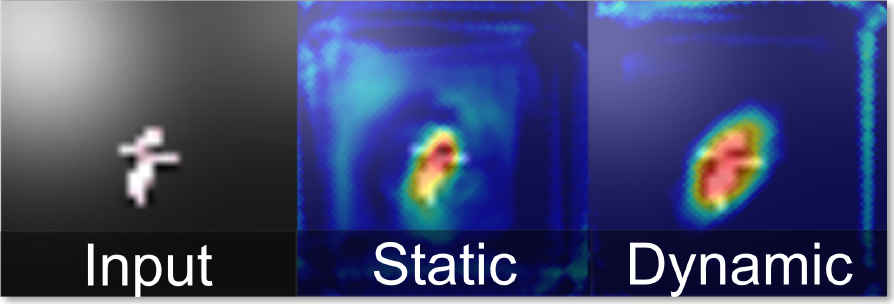}
        \subcaption{\small{Spiky object 1}}
    \end{minipage}
    \begin{minipage}[tb]{0.49\linewidth}
        \centering
        \includegraphics[keepaspectratio, width=\linewidth]{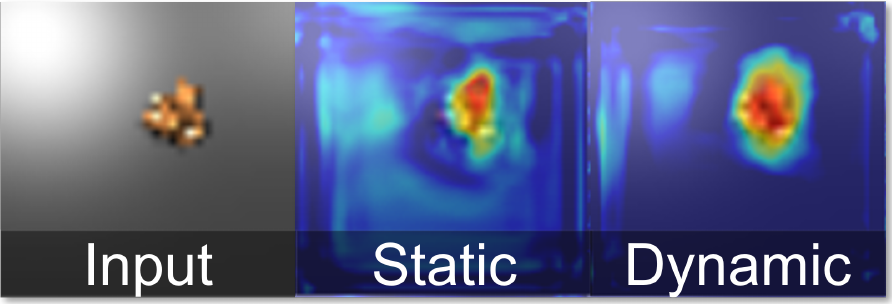}
        \subcaption{\small{Spiky object 2}}
    \end{minipage}
    \caption{\small{$Q$ values generated from the three models of the static grasping, moving, and dynamic grasping modules for input images}}
    \figlab{predicted}
\end{figure}
\figref{predicted} shows the heightmap image inputs for different T-shaped (a), (b) and spiny (c), (d) objects and the respective outputs $Q_{\rm{gs}}$ and $Q_{\rm{gd}}$ of the model after training with the PP.
Specifically, the figures show input images taken directly above the target objects (Input), static grasp score maps (Static) in which the outputs of the static grasping module are superimposed on the input images, and dynamic grasp score maps (Dynamic) in which the outputs of the dynamic grasping module are superimposed on the input images.

In the case of (a), the static grasp score map indicates that the upper part of the T-shape should be grasped, but in the dynamic grasp score map, there is an equivalent pixel value area in the lower part of the T-shape.
This is because the succeeding fingertip may be conveniently caught in the upper part of the T-shape and succeed in grasping when approaching from the left side of the image.
This indicates that the robot itself may have acquired knowledge of mobile grasping that could not be obtained by static grasp learning alone, which is one of the effects of combined learning for mobile grasping.
In the case of (b), there was no such tendency because the succeeding fingertip was in a posture that was not caught.
For the T-shape, the robot could predict the possible grasping position with a higher degree of confidence than with the static grasping module alone by considering the movement.

\kiyokawa{In cases (c) and (d), similar pixel value areas in the score maps may have expanded due to the task's uniform ease or difficulty. Our future work will explore whether larger datasets and additional training steps yield more stable grasp score maps relative to object shapes.}
However, the average success rate of mobile grasping by the PP at the three different speeds was 80\% for both the spiky object 1 and 2, which is almost the same as the overall average of 83\% for PP as shown in~\tabref{res_v}, suggesting that all of the positions were graspable shapes.

Furthermore, as a result of predicting the residual velocity using the PP in 1000 different randomly configured environments, the mean $\pm$ standard deviation values for $v=0.10, 0.15, 0.20$ were $0.004\pm0.0016$, $0.003\pm0.0013$, $0.002\pm0.0016$.
This suggests that the PP can predict relatively small residual velocity values for the actual value of velocity given by the user input and that the velocity can be finely adjusted.

\section{Evaluating Real-World Mobile Grasping}
\begin{figure}[tb]
    \centering
    \small
    \begin{minipage}[tb]{\linewidth}
        \centering
        \includegraphics[keepaspectratio, width=\linewidth]{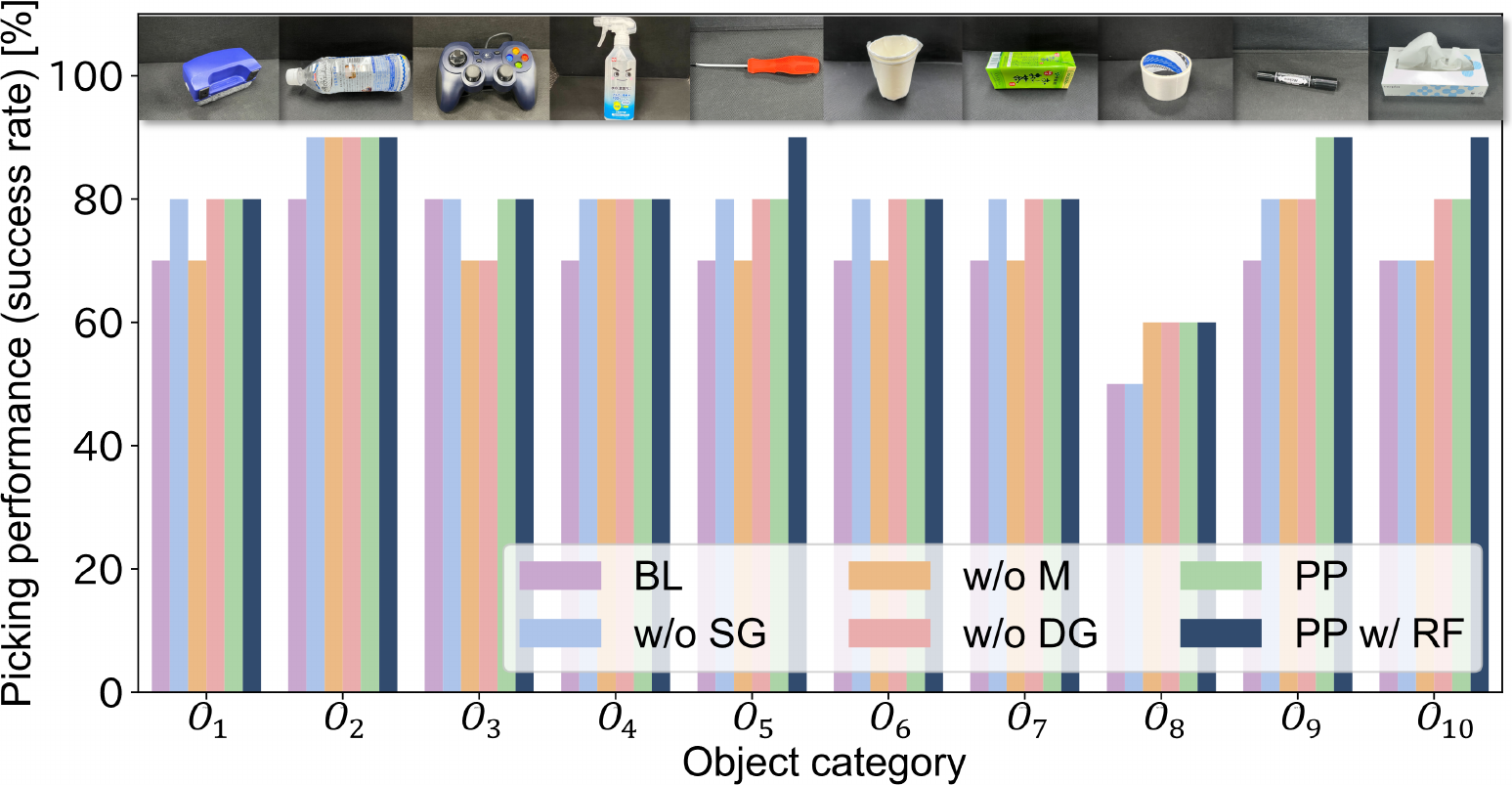}
        \subcaption{\small{$v=0.10$}}
    \end{minipage}
    \begin{minipage}[tb]{\linewidth}
        \centering
        \includegraphics[keepaspectratio, width=\linewidth]{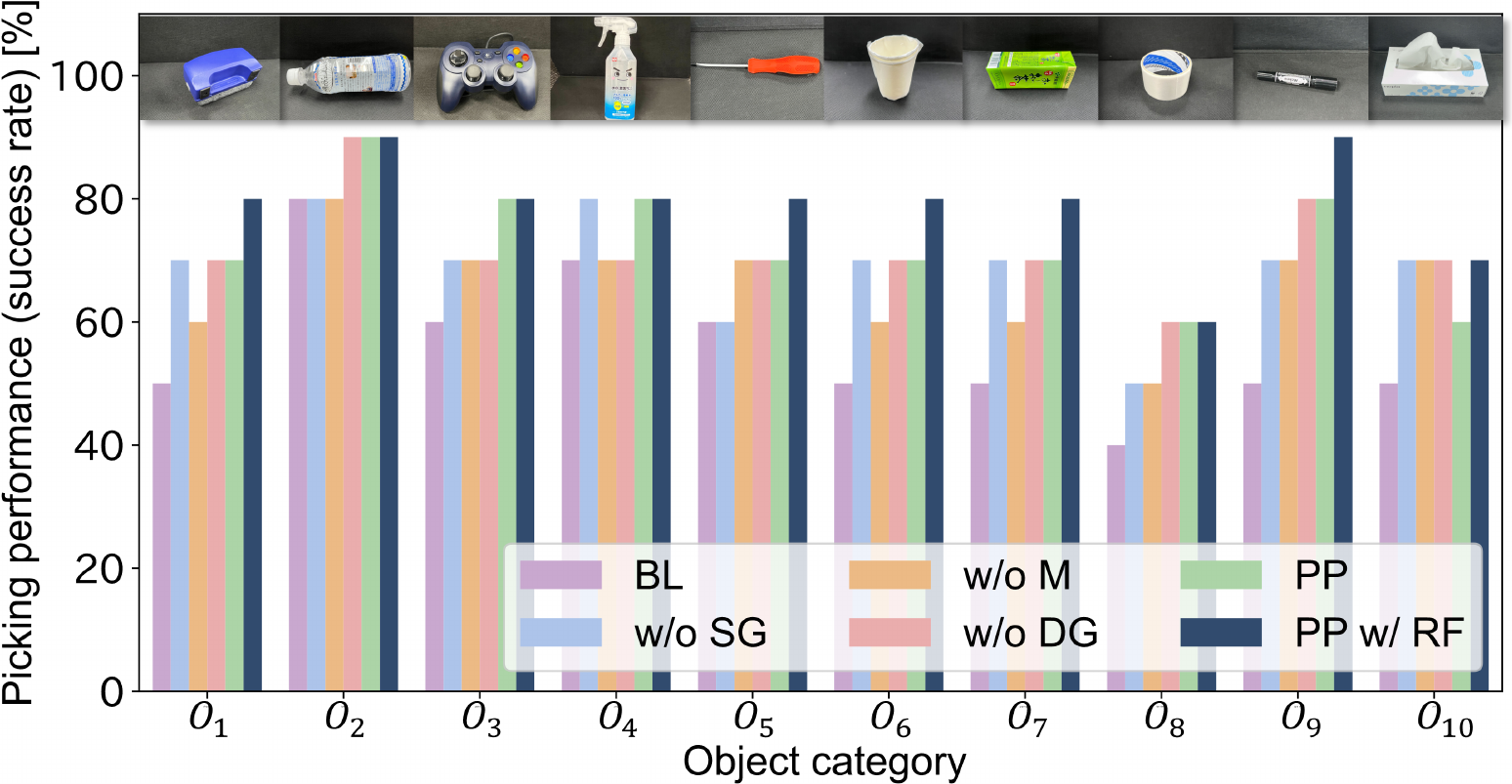}
        \subcaption{\small{$v=0.15$}}
    \end{minipage}
    \begin{minipage}[tb]{\linewidth}
        \centering
        \includegraphics[keepaspectratio, width=\linewidth]{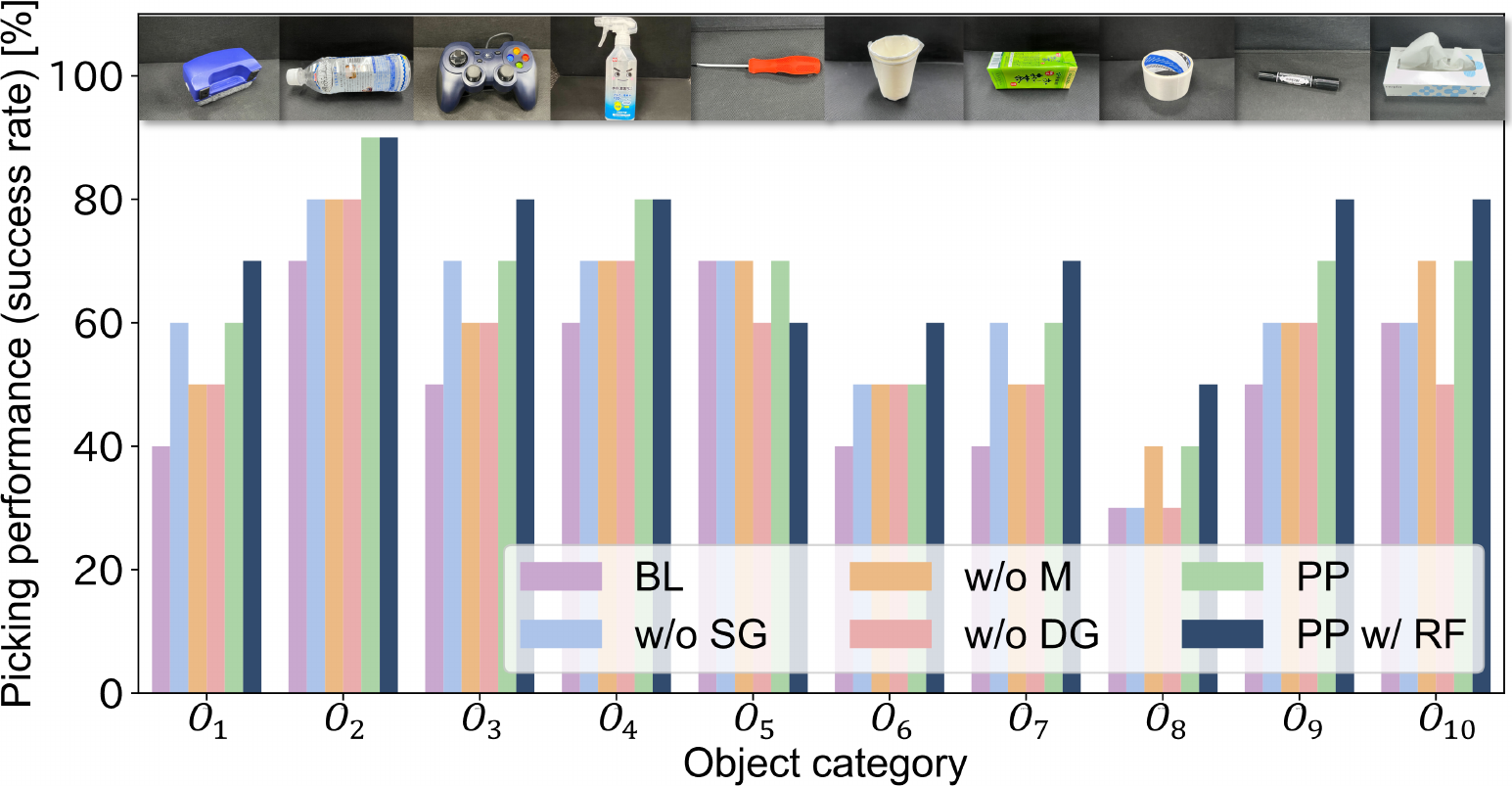}
        \subcaption{\small{$v=0.20$}}
    \end{minipage}
    \caption{\small{Performance of real-world mobile grasping. The graphs show success rates for (a) $v=0.10$, (b) $v=0.15$, and (c) $v=0.20$.}}
    \figlab{real}
\end{figure}
\begin{table}[tb]
    \centering
    \small
    \begin{threeparttable}
        \caption{\small{Mobile grasping success rate in the real world [\%]}}
        \tablab{success_real}
        \begin{tabular}{cp{12mm}p{12mm}p{12mm}p{12mm}p{12mm}} \toprule
            \multicolumn{1}{c}{$v$[m/s]} & \multicolumn{1}{c}{~~~BL~~} & \multicolumn{1}{c}{~w/o SG~} & \multicolumn{1}{c}{~~w/o M~~} & \multicolumn{1}{c}{~w/o DG~} & \multicolumn{1}{c}{~~~PP~~~} \\ \midrule
            0.10 & \multicolumn{1}{r}{70} & \multicolumn{1}{r}{73} & \multicolumn{1}{r}{78} & \multicolumn{1}{r}{77} & \multicolumn{1}{r}{\textbf{80}} \\
            0.15 & \multicolumn{1}{r}{56} & \multicolumn{1}{r}{66} & \multicolumn{1}{r}{72} & \multicolumn{1}{r}{69} & \multicolumn{1}{r}{\textbf{73}} \\
            0.20 & \multicolumn{1}{r}{51} & \multicolumn{1}{r}{60} & \multicolumn{1}{r}{56} & \multicolumn{1}{r}{61} & \multicolumn{1}{r}{\textbf{66}} \\ \midrule
            Mean & \multicolumn{1}{r}{59}	& \multicolumn{1}{r}{66} & \multicolumn{1}{r}{69} & \multicolumn{1}{r}{69} & \multicolumn{1}{r}{\textbf{73}} \\ \bottomrule
        \end{tabular}
    \end{threeparttable}
\end{table}
\begin{table}[tb]
    \centering
    \small
    \begin{threeparttable}
        \caption{\small{Comparison on efficiency of mobile grasping actions}}
        \tablab{time_real}
        \begin{tabular}{lp{20mm}p{20mm}p{20mm}p{20mm}} \toprule
            & \multicolumn{4}{c}{$v$[m/s]} \\ \cmidrule(r){2-5}
            & \multicolumn{1}{c}{~~~0.00~~~} & \multicolumn{1}{c}{~~~0.10~~~} & \multicolumn{1}{c}{~~~0.15~~~} & \multicolumn{1}{c}{~~~0.20~~~} \\ \midrule
            Mean time~[s]~~ & \multicolumn{1}{r}{~~~33.7~~~} & \multicolumn{1}{r}{~~~22.5~~~} & \multicolumn{1}{r}{~~~19.7~~~} & \multicolumn{1}{r}{~~~\textbf{17.7}~~~} \\
            MPPH~~ & \multicolumn{1}{r}{~~~106~~~} & \multicolumn{1}{r}{~~~160~~~} & \multicolumn{1}{r}{~~~182~~~} & \multicolumn{1}{r}{~~~\textbf{203}~~~} \\ \bottomrule
        \end{tabular}
    \end{threeparttable}
\end{table}
To verify the grasping success rate and collection time of the target objects in a real-world environment, our real-world experiments used the models learned in the simulations.
To further validate real-world fine-tuning (RF), we trained the models via trial and error for 150 steps for the five different object shapes shown in~\figref{obj}~(b); thus, we added another method named PP w/ RF in this experiment.
We used 10 different objects, $O_j (j \in \{1,2,\ldots,10\})$, as shown in~\figref{real}.
We used an HSR to implement the mobile grasping actions, and an RGB-D sensor Kinect v2 to obtain the input image.

At the beginning of the experiments, one target object was randomly placed in a workspace within the sensor's sensing range such that the distance between the positions of the robot and target object was 1.0–1.5m.
The robot started moving from a predetermined position indicated by the white frame on the floor, as shown in~\figref{cs}.
Subsequently, the mobile grasping actions generated by each method were executed.
Our experiment performed mobile grasping actions 10 times for each target object, and calculated the grasping success rate [\%].
After lifting the gripper, the target object was visually checked to see if it had fallen from the gripper, and if it had not fallen, the grasp was considered successful.
Object collection time was measured as the time taken for the robot to start moving laterally, grasping, and lifting the closed gripper.
The measurement results of the success rate for each target object, the grasping success rate for each set velocity, and the object collection time are shown in \figref{real}, \tabref{success_real}, and \tabref{time_real}, respectively.

\subsection{Grasping Success Rate Comparisons}
Looking at the success rate for each target object shown in~\figref{real}, there is no significant performance difference between the methods for $v=0.10$ and $v=0.15$.
However, for $v=0.20$, there is a noticeable difference in the success rate.
Specifically, the success rates for long objects such as the plastic bottle ($O_2$), spray bottle ($O_4$), screwdriver ($O_5$), paper pack ($O_7$), pencil ($O_9$), and tissue box ($O_{10}$) were relatively high at over 70\%, whereas the success rate for relatively short objects, such as the whiteboard cleaner ($O_1$), gamepad ($O_3$), paper cup ($O_6$), and tape ($O_8$), dropped significantly.
This is probably because the tolerable moving error is small because the object shape is short, and the timing at which it can be grasped is less than that for long objects, which cannot accommodate higher velocities.
Similarly, for the long objects ($O_2$, $O_4$, $O_5$, $O_7$, $O_9$, $O_{10}$), there was no decrease in the success rate for either method at any velocity.
This suggests that long objects are relatively easy to grasp.

\kiyokawa{The lowest success rate was observed for cylindrical object $O_8$ across all velocities and methods. Successful grasps for $O_8$ occurred only when one finger was inside the cylinder and the other outside. The cylinder's internal diameter was narrower, complicating the grasp compared to other objects.}

\kiyokawa{Trajectory planning and online control technologies}~\cite{Colombo2019,Zimmermann2021,Thakar2022tase,Limerick2023} \kiyokawa{offer promising solutions for enhancing motion stability and robustness across various object shapes. Our future work will explore integrating the proposed grasping planner with its trajectory planner and online controllers.}

Considering the success rate for each velocity setting shown in~\tabref{success_real}, the proposed method was the highest at all velocities.
It is noteworthy that compared to the simulation results, the success rate in the real-world experiments decreased significantly as $v$ increased.
\kiyokawa{In fact, when the velocity was increased from 0.10 to 0.20, the drop in the grasping success rate was 5\% for BL, 3\% for w/o SG, 6\% for w/o M, 3\% for w/o DG, and 1\% for PP in the simulation case, whereas in the real-world experiment, they were 19\%, 13\%, 12\%, 9\%, and 14\%, respectively.
This may be due to the fact that the actual robot's moving error was not taken into account during the training, and the estimated grasping position and orientation may not have been approached correctly.

However, the proposed method showed the highest rate over the comparative methods, both in the simulation and real-world experiments, proving the possibility of the proposed method performing mobile grasping over a wide range of velocities.}
Furthermore, the success rate of w/o M did not drop as sharply as that of BL, even when $v$ increased, despite the fact that this method does not have a moving module.
This may be due to the fact that the dynamic grasping module may also play the role of the moving module, which may be able to accommodate some errors.
The results of PP w/ RF outpeforms those of any other methods, suggesting the results of this experiment also demonstrated the effectiveness of additional learning in the real world.

\kiyokawa{While the use of randomly colored workspaces in the simulation environment improved performance, our future work will address the sim2real problem. Possible solutions include real2sim}~\cite{Rao2020} \kiyokawa{and randomized to-canonical adaptation}~\cite{James2019} \kiyokawa{networks. For novel object grasping, transferring grasp inference for the trained object or object registered in database to the real-world target based on their similarity}~\cite{Chen2022} \kiyokawa{may be effective to enhance the performance.}

\subsection{Pick-and-Place Efficiency}
\begin{figure}[tb]
    \centering
    \small
    \begin{minipage}[tb]{0.49\linewidth}
        \centering
        \includegraphics[keepaspectratio, width=\linewidth]{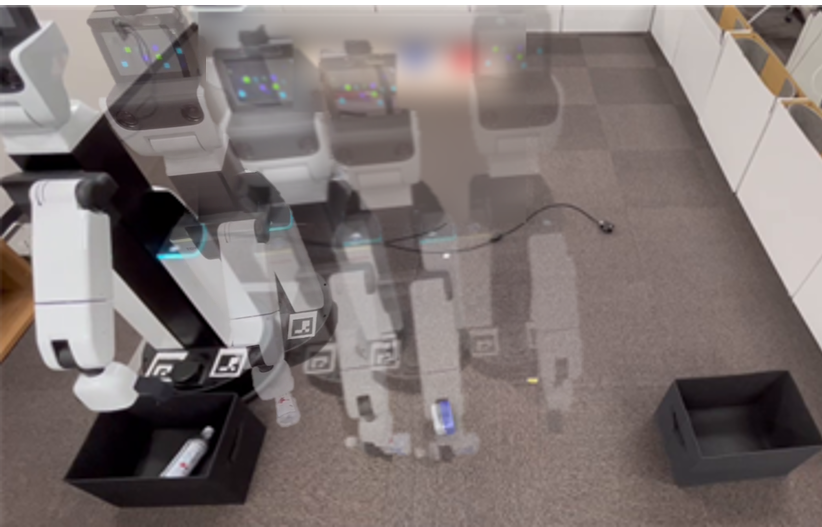}
        \subcaption{\small{First action for $O_2$}}
    \end{minipage}
    \begin{minipage}[tb]{0.49\linewidth}
        \centering
        \includegraphics[keepaspectratio, width=\linewidth]{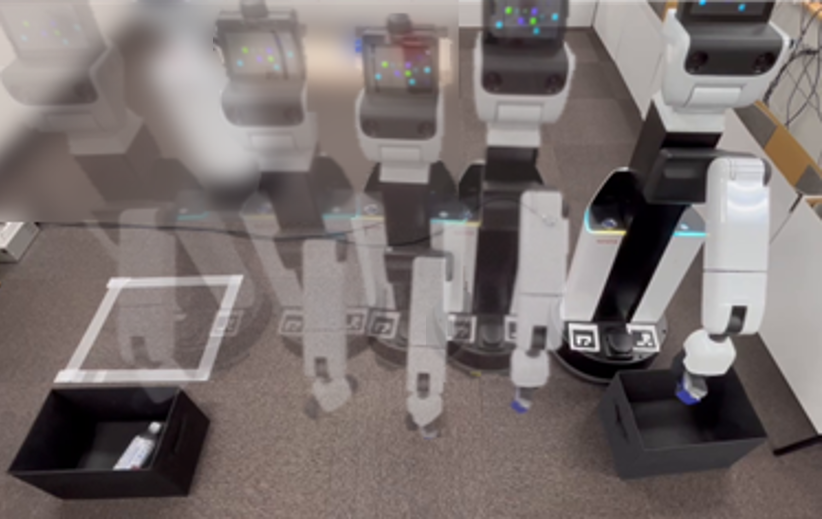}
        \subcaption{\small{Second action for $O_1$}}
    \end{minipage}
    \caption{\small{Multiple mobile grasping actions applied for two different objects, (a) $O_2$ and (b) $O_1$.}}
    \figlab{multi_mg}
\end{figure}
\begin{figure}[tb]
    \centering
    \small
    \includegraphics[keepaspectratio, width=\linewidth]{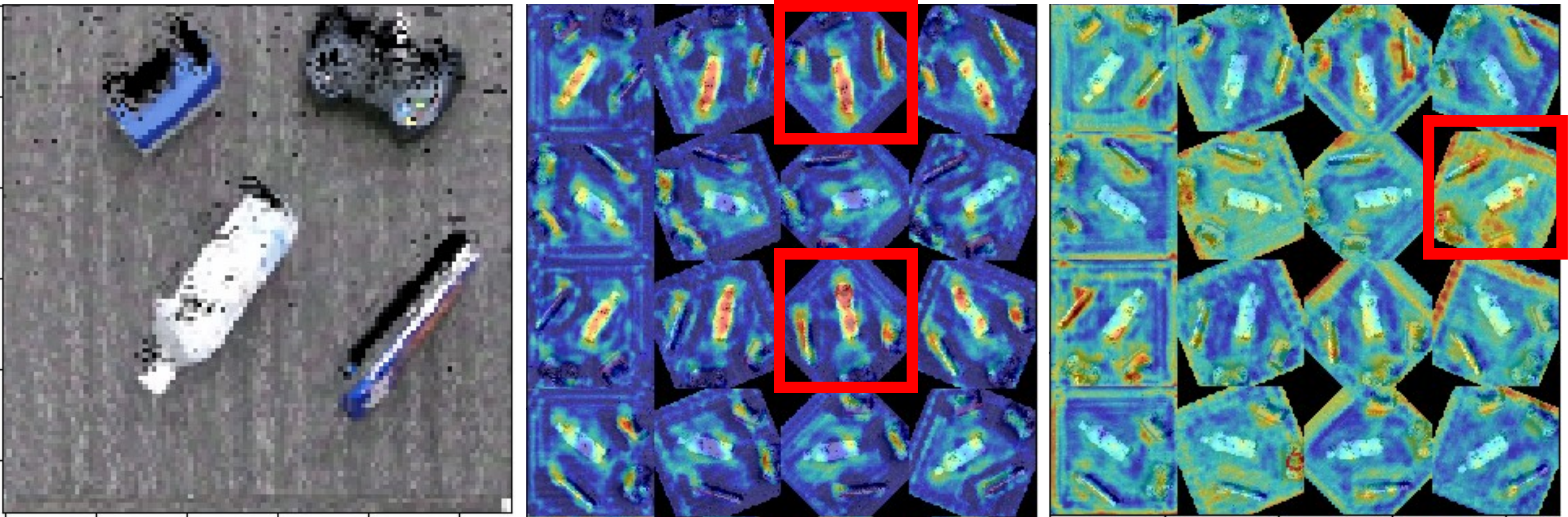}
    \caption{\small{Prediction results for multiple objects placed in the workspace. From left to right, the figure shows the input image, static grasp prediction, and dynamic grasp prediction.}}
    \figlab{mo}
\end{figure}
We compared the difference in object collection time between PP for tidying up by performing mobile grasps and tidying up by sequentially stopping in front of the target object.
The velocity setting verified in this experiment is $v \in \{0.00, 0.10, 0.15, 0.20\}$[m/s].
The sequential stopping is expressed as $v=0.00$.

\tabref{time_real} shows the measurement results of object collection time.
The results suggest that even with the slowest mobile grasp, $v=0.10$, the grasp trial ended more than 10 s earlier than the sequential stopping.
The object could only be picked 106 times/hour (Mean Picks Per Hour (MPPH)) with the sequential stopping grasping, but the picking speed improved to 203 times/hour with the mobile grasping.
With mobile grasping, the mobile manipulator could pick the object 203 times per hour, achieving a grasping success rate of 76\%.
Mobile grasping has approximately twice the MPPH as sequential stopping grasping, and the PP allows mobile grasping with a high success rate and is expected to improve the efficiency of tidy-up tasks.

As shown in~\figref{multi_mg}, even in an environment where multiple objects are placed, the grasp point and orientation of the objects can be predicted, and multiple mobile grasping actions can be applied to the environment.

\figref{mo} shows another example of grasp probability prediction result for multiple objects placed in the workspace. From left to right, the figure shows the input image, static grasp prediction, and dynamic grasp prediction.
The target objects include objects $O_1$, $O_2$, $O_3$ used in the experiments and another unknown object (long pen).

The red-framed area of the static grasp prediction result shows that for long objects such as $O_2$ and the long pen, the case when the robot's moving direction and the direction connecting the two fingertip positions is parallel have a high probability of grasp success. However, the dynamic grasp prediction result indicates that a perpendicular orientation of the fingers to the moving direction has a higher probability of grasp success, as it is more robust against grasp uncertainty in that direction. This dynamic grasping module's prediction suggests that the proposed method can also predict the grasp success probability for multiple target objects correctly.

Our future work will focus on experiments in a more cluttered environment and tackle the tidy-up scheduling problem~\cite{Yan2021,Su2023} to achieve both a high grasping success rate and high efficiency.
Leveraging the predicted successful grasp probability to determine the order of objects to be grasped may be a possible approach in this research direction.

\section{Conclusion}
\kiyokawa{This study aims to enable a commercial off-the-shelf robot to perform mobile grasping by fine-tuning the timing and pose of the grasp. A self-supervised learning framework was introduced to quickly equip a mobile manipulator with object-agnostic mobile grasping capabilities. To address data sparsity due to the complexity of mobile grasping, this study introduced methods of action primitivization and step-by-step learning for divided action. The framework comprises three essential modules: a moving module adjusting velocity, a static grasping module for stationary objects, and a dynamic grasping module for moving objects, each employing fully convolutional neural network (FCN) models. A two-stage grasp learning approach was proposed to integrate the three action-primitive-specific FCN models.

Simulation and real-world experiments demonstrated that the proposed method outperformed comparative methods across various velocity settings, showing higher grasping accuracy and pick-and-place efficiency. Randomizing object shapes and environments in simulations effectively achieved generalizable mobile grasping.

Future work will address planning problems to optimize the order and trajectory for collecting multiple tidy-up targets to achieve higher efficiency. Furthermore, the impact of larger datasets, additional training steps, sim2real approaches, and similarity-based grasping ability transfer will be investigated to enhance grasping performance for diverse objects, including novel ones.}

\bibliographystyle{IEEEtran}
\footnotesize
\bibliography{reference}

\end{document}